\documentclass{article}

\usepackage{microtype}
\usepackage{graphicx}
\usepackage{booktabs} 

\usepackage{hyperref}

\usepackage[accepted]{icml2024}

\usepackage{amsmath}
\usepackage{amssymb}
\usepackage{mathtools}
\usepackage{amsthm}
\usepackage{bbm}
\usepackage{subcaption}
\usepackage{enumerate}

\newdimen\newcaptionboxwid
\makeatletter
\long\def\@makecaption#1#2{%
  \vskip 10pt
  \baselineskip 11pt
  \setbox\@tempboxa\hbox{#1. #2}%
  \ifdim \wd\@tempboxa >\hsize
    \sbox{\newcaptionbox}{\small\sl #1.~}%
    \newcaptionboxwid=\wd\newcaptionbox
    \usebox\newcaptionbox {\footnotesize #2}
  \else
    \centerline{{\small\sl #1.} {\small #2}}
  \fi
}
\makeatother

\usepackage[capitalize,noabbrev]{cleveref}

\theoremstyle{plain}

\theoremstyle{definition}

\theoremstyle{remark}

\usepackage[textsize=tiny]{todonotes}

\definecolor{blue}{HTML}{4878D0}
\definecolor{red}{HTML}{D65F5F}
\definecolor{green}{HTML}{6ACC64}
\definecolor{orange}{HTML}{EE854A}
\definecolor{grey}{HTML}{797979}
\definecolor{lightgrey}{HTML}{adb3b7}
\definecolor{purple}{HTML}{7D3382}
\newcommand{\res}[2]{$#1 \color{lightgrey} \pm \small#2$}

\usetikzlibrary{positioning}

\icmltitlerunning{Pretrained Visual Uncertainties}

\begin{document}

\twocolumn[
\icmltitle{Pretrained Visual Uncertainties}



\icmlsetsymbol{equal}{*}

\begin{icmlauthorlist}
\icmlauthor{Michael Kirchhof}{tub}
\icmlauthor{Mark Collier}{google}
\icmlauthor{Seong Joon Oh}{ai}
\icmlauthor{Enkelejda Kasneci}{muc}
\end{icmlauthorlist}

\icmlaffiliation{tub}{University of Tübingen, Germany}
\icmlaffiliation{muc}{TUM University, Munich, Germany}
\icmlaffiliation{ai}{University of Tübingen, Tübingen AI Center, Germany}
\icmlaffiliation{google}{Google Research, Switzerland}

\icmlcorrespondingauthor{Michael Kirchhof}{michael dot kirchhof at uni dash tuebingen dot de}

\icmlkeywords{Uncertainty Quantification, Pretrain, Uncertainties, ImageNet, ViT, Vision Transformer, Machine Learning, ICML}

\vskip 0.3in
]



\printAffiliationsAndNotice{\icmlEqualContribution} 

\begin{abstract}

Accurate uncertainty estimation is vital to trustworthy machine learning, yet uncertainties typically have to be learned for each task anew. This work introduces the first pretrained uncertainty modules for vision models. Similar to standard pretraining this enables the zero-shot transfer of uncertainties learned on a large pretraining dataset to specialized downstream datasets. We enable our large-scale pretraining on ImageNet-21k by solving a gradient conflict in previous uncertainty modules and accelerating the training by up to 180x. We find that the pretrained uncertainties generalize to unseen datasets. In scrutinizing the learned uncertainties, we find that they capture aleatoric uncertainty, disentangled from epistemic components. We demonstrate that this enables safe retrieval and uncertainty-aware dataset visualization. To encourage applications to further problems and domains, we release all pretrained checkpoints and code under \url{https://github.com/mkirchhof/url}.
\end{abstract}

\section{Introduction}

With every prediction comes the risk of an error. Uncertainty estimates quantify this expected error in order to defer predictions and catch errors before they happen, a key requirement for trustworthy machine learning \cite{mucsanyi2023trustworthy}. 
Uncertainty quantification has seen tremendous advances in recent years, bringing principled methods such as Gaussian processes \cite{liu2020simple} and probabilistic embeddings \cite{oh2018modeling,kirchhof2023probabilistic,pmlr-v202-kim23g,nakamura2023representation} to large-scale computer vision \cite{tran2022plex,dehghani2023scaling,collier2023massively}. Recent benchmarks reveal that they excel at their metrics and are ready for application \cite{galil2023a,galil2023what}.
However, there is a lack of widespread adoption of uncertainty methods by practitioners. The hurdle is that modern uncertainty quantification methods can be complex, making them difficult to implement and increasing the inference costs. So what if uncertainty estimates were as easy to access as being shipped along with every pretrained model?

We seek a method that is simple to implement and train, even on large scales, and most importantly does not interfere with the main objective of a practitioner's model. One group of recent methods stands out on these terms: Feed-forward uncertainties \cite{cui2023learning,kirchhof2022non,Yoo_2019_CVPR,oh2018modeling}. They an auxiliary uncertainty head to a deep network that is evaluated alongside every forward pass. Not only is this cheap and simple to implement, it was also recently found that its uncertainty estimates transfer well \cite{kirchhof2023url}.

\begin{figure}[t]
    \centering
    \begin{tikzpicture}
    \node[anchor=south west,inner sep=0] (image) at (0,0) {\includegraphics{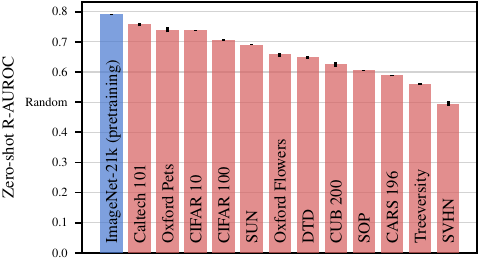}};
    \begin{scope}[x={(image.south east)},y={(image.north west)}]
        \node[anchor=north,inner sep=1mm] (text) at (0.613,0.02) {\footnotesize \textcolor{red}{unseen datasets}};
        \draw[red] (text.west) -- (0.293, -0.028);
        \draw[red] (text.east) -- (0.934, -0.028);
    \end{scope}
    \end{tikzpicture} \vspace{-2mm}
    \caption{Our pretrained uncertainties generalize to unseen datasets. The R-AUROC measures the quality of uncertainty estimates on zero-shot datasets, see \cref{sec:quant}.}
    \label{fig:generalize}
\end{figure}
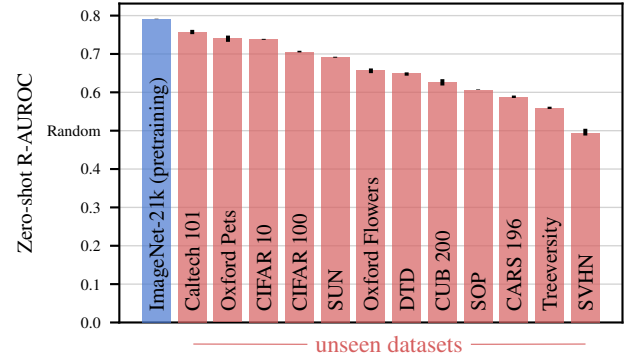

Our work solves a remaining gradient conflict of these feed-forward uncertainties to guarantee non-interference with the main objective. We also implement massive caching in our pretraining pipeline, reducing the train time by a factor of up to 180x. This enables us to scale up both the pretraining dataset to ImageNet-21k Winter-2021 (ImageNet-21k-W) \cite{deng2009imagenet} and the vision backbone to large Vision Transformers \cite{vit}. 

\cref{fig:generalize} shows that our pretrained uncertainties now transfer beyond the train dataset to unseen datasets, outperforming previous zero-shot uncertainties \cite{kirchhof2023url}. We find that the learned uncertainties are generalizable notions of aleatoric uncertainty disentangled from epistemic uncertainty. This enables multiple use-cases: Beyond error prediction, we showcase novel applications like safe retrieval and uncertainty-aware dataset visualization. To facilitate widespread adoption, we release checkpoints for our pretrained uncertainties along with efficient code to pretrain them for arbitrary model architectures. 

In summary, our contributions are:
\begin{itemize}
    \item We develop a method which learns pretrained uncertainties that transfer zero-shot.
    \item This is based on fixing a gradient conflict in previous feed-forward uncertainties and speeding up the training by 180x, enabling large-scale pretraining (\cref{sec:enhance}).
    \item Our uncertainties represent aleatoric uncertainty, disentangled from epistemic uncertainty (\cref{sec:understand}).
    \item We apply the uncertainties to improve the reliability of retrieval and to aid data visualization (\cref{sec:applications}).
\end{itemize}

\section{Related Work}

\textbf{Large-scale uncertainty quantification.} Uncertainty quantification has recently been scaled rapidly. Within one year, the largest vision models capable to perform uncertainty quantification grew from 1B \cite{tran2022plex} to 22B parameters \cite{dehghani2023scaling}. Benchmarks have increased accordingly. Previous surveys on out-of-distribution detection of 32x32 sized images \cite{ovadia2019can} have been scaled to real-world images \cite{galil2023a} and to several additional tasks \cite{galil2023what}. One such task is the uncertainty-aware representation learning (URL) benchmark that pretrains an uncertainty estimator and then tests zero-shot uncertainties on unseen datasets \cite{kirchhof2023url}. Our work sets a new state-of-the-art on this task. In particular, we provide pretrained uncertainty modules for large computer vision models, independent from the classes or specific task of a dataset.

\textbf{Feed-forward uncertainties.} State-of-the-art models attempt to give such transferable uncertainties by moving away from classifier-layer uncertainties and towards uncertainties in the representation space \cite{collier2023massively}. This approach falls under the category of feed-forward or deterministic uncertainties \cite{pmlr-v162-postels22a}. They have a specialized uncertainty module that outputs predicted uncertainties during the forward pass of the model at minimal computational costs, which enables scaling. A variational take on this are probabilistic embeddings \cite{oh2018modeling,chun2023improved,pmlr-v202-kim23g,nakamura2023representation} that output a variance estimate to give  a distribution of possible representations instead of just one. This has recently been proven to recover the aleatoric uncertainty of the true posterior \cite{kirchhof2023probabilistic} and improve retrieval performance \cite{karpukhin2022probabilistic}. As opposed to such indirect approaches, a second group of feed-forward approaches makes uncertainty quantification a direct regression task \cite{Yoo_2019_CVPR,cui2023learning,lahlou2023deup,laves2020well}. In initial experiments, we found this direct approach to scale better. We use it as a starting point to develop our pretrained uncertainties in the next section, overcoming some remaining challenges to enable scaling.

\section{Developing Pretrained Uncertainties} \label{sec:dev}

In this section we set out the desired properties of our pretrained uncertainties and then extend a popular feed-forward uncertainty method to satisfy these properties in the large-scale pretraining case. 

\subsection{Basic Principles} \label{sec:principles}
We develop pretrained uncertainties from basic principles for scalability and ease of use, in order of importance:
\begin{enumerate}[(i)]
\item \textbf{Non-interference with primary task}. Adding pretrained uncertainties to a model should not worsen the performance of the pretrained model's primary objective, e.g., its accuracy.

\item \textbf{Generalization}. The predicted uncertainty estimates should reflect general forms of uncertainty that transfer to unseen datasets and tasks beyond the pretraining data and task.

\item \textbf{Flexible adjustment}. The uncertainties should be general enough to adapt to new tasks and/or datasets that downstream practitioners might introduce.

\item \textbf{Minimal overhead}. Providing uncertainties add only minimal runtime and memory usage to the main task prediction model.

\item \textbf{Scalable optimization}. Training should converge stably to ensure scalability to large pretraining corpora.
\end{enumerate}

In summary, we seek a \emph{download and forget} approach that requires minimal interventions from practitioners.

\begin{figure}[t]
    \centering
    \includegraphics[trim=0cm 8.8cm 17cm 0cm, clip, scale=0.7]{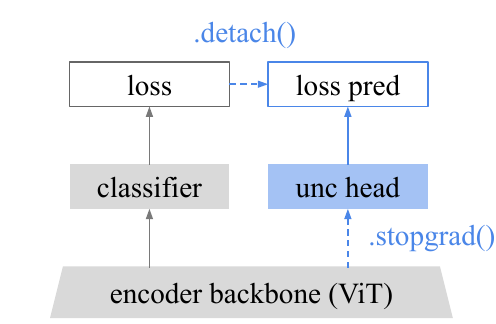}
    \caption{Pretrained uncertainties are returned by an auxiliary head (blue) that is trained to predict the classification loss of each image.}
    \label{fig:architecture}
\end{figure}

\begin{figure*}[t]
\centering
\begin{subfigure}[b]{0.43\linewidth}
  \centering
  \includegraphics{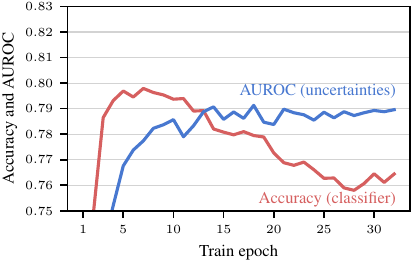}\vspace{0.5mm}
  \caption{\footnotesize Vanilla Loss Prediction}
  \label{fig:sub1}
\end{subfigure}\hfill%
\begin{subfigure}[b]{0.43\linewidth}
  \centering
  \includegraphics{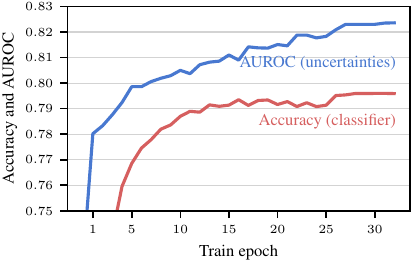}\vspace{0.5mm}
  \caption{\footnotesize Ours: Loss Prediction + Stopgrad}
  \label{fig:sub2}
\end{subfigure}
  \begin{tikzpicture}[overlay, remember picture]
    \draw[->, line width=1pt] (-9.4,3.2) -- (-8.0,3.2);
    \node[above, text centered] at (-8.7,3.2) {\footnotesize + stopgrad};
  \end{tikzpicture} 
  
  \vspace{-1mm}
\caption{(a) The uncertainty and classification heads of Loss Prediction are in conflict. We solve this in (b) by adding a stopgrad. It ensures that the uncertainty head's gradients do not interfere with those of the classifier head, stabilizing the performance of both. The uncertainty and classifier heads were finetuned on ImageNet-1k on a pretrained (but unfrozen) ViT-Base backbone.}
\label{fig:test}
\end{figure*}

\subsection{Recap: Loss Prediction} \label{sec:losspred}
We now introduce a simple yet general uncertainty method that we build our method upon. From a decision theory perspective, giving an uncertainty estimate means predicting how wrong one thinks one's estimate is. The key is that any task's level of wrongness is defined by its loss $\mathcal{L}_\text{task}$. So in loss prediction \cite{Yoo_2019_CVPR,cui2023learning,lahlou2023deup,laves2020well}, the model has an additional module $u$ that predicts the model's own loss at each of its predictions. This is learned via a $L_2$ loss between $u$ and the (gradient-detached, det.) main task loss $\mathcal{L}_\text{task}^\text{det.}(y, f(x))$ at every sample $(x, y)$. This is trained along with the main task \cite{kirchhof2023url}, yielding the combined objective $\mathcal{L}$:
\begin{align}
    \mathcal{L} = \mathcal{L}_\text{task}(y, f(x)) + (u(e(x)) - \mathcal{L}_\text{task}^\text{det.}(y, f(x)))^2\,.
\end{align}
The uncertainty module $u$ is implemented as a small MLP head $u(e(x))$ on top of the model representations $e(x)$. This makes it cheap to compute during the forward pass, fulfilling the minimal overhead principle (iv). Loss prediction's uncertainties also adapt to any loss, fulfilling the flexibility principle (iii), and transfer well \cite{kirchhof2023url}, fulfilling the generalization principle (ii).

Yet, loss prediction's implementation has a limitation: \cref{fig:sub1} depicts a conflict between the uncertainty and the classification task. Their gradients interact negatively with one another, deteriorating the joint backbone and violating the non-interference principle (i). To resolve it, the current implementation stops early, roughly at epoch 12 in the plot. However, this early stopping is at odds with the scalability principle (v). Below, we fix these issues.

\subsection{Enhancing Loss Prediction} \label{sec:enhance}

We introduce four changes to the above loss prediction:

\textbf{1. Stopgrad.} As visualized in \cref{fig:architecture}, we add a stopgrad behind the uncertainty module. This prevents its gradients from flowing to the backbone and interfering with the classification head. This strictly ensures the non-interference principle (i), and, indeed, the training now converges robustly, see \cref{fig:sub2}. This way, uncertainties can be trained in parallel to the main task, as opposed to only in post-hoc.

\textbf{2. No early stopping.} With the gradient conflict resolved, there is no more need for early stopping. The uncertainty head converges to its maximum at the end of the training in \cref{fig:sub2}, making the training scalable as per principle (v).

\textbf{3. Cache everything.} Since the classification head and backbone are now independent from the uncertainty head, we pretrain and then freeze them before training the uncertainty module. Only the uncertainty objective remains: 
\begin{align}
    \mathcal{L} = (u(e(x)) - \mathcal{L}_\text{task}^\text{det.}(y, f(x)))^2
\end{align}
This can be optimized efficiently: The uncertainty module uses only the representations $e(x)$ as inputs, and, likewise, the task loss depends only on them via $f(x) = c(e(x))$, where $c$ is the classifier layer. So, we do not need to load the images $x$ or run them through the backbone, but can cache the representations $e(x)$ of the whole training process once (all epochs, including random augmentations). When learning the uncertainty module on top of a pretrained model, this increases the train speed by a factor of 180x and reduces the memory usage so far that we can pretrain uncertainties even for large models on single GPUs (or even CPUs). This paves the way for scalability: After caching the representations once, training the uncertainty module of a ViT-Large for seven ImageNet-21k-W epochs takes 2:26 hours on a single V100 GPU as opposed to 18 days with the standard loss prediction implementation.

\textbf{4. Scale-free uncertainties.} With the current $L_2$ loss, the uncertainty module is trained to match the scale of the pretraining loss. However, a downstream user might switch to a different loss on a different scale, which would introduce destructive gradients during finetuning. Thus, we switch to the ranking-based objective of \citet{Yoo_2019_CVPR}:
\begin{align}
    & \mathcal{L} = \max(0, \mathbbm{1}_\mathcal{L} \cdot (u(e(x_1)) - u(e(x_2)) + m)), \\
    & \text{s.t. } \hspace{-0.2mm}\mathbbm{1}_\mathcal{L} \hspace{-0.5mm}:=\hspace{-0.4mm} \begin{cases}
        +1 & \hspace{-3mm}\text{, if } \mathcal{L}_\text{task}^\text{det.}(y_1, f(x_1)) \hspace{-0.4mm}>\hspace{-0.4mm} \mathcal{L}_\text{task}^\text{det.}(y_2, f(x_2)) \\
        -1 & \hspace{-3mm}\text{, else} 
    \end{cases} \hspace{-3mm}
\end{align}
For every pair of images $x_1$ and $x_2$, the indicator function compares which image has the higher primary task loss $\mathcal{L}_\text{task}^\text{det.}$. Then, the uncertainty $u$ of that sample is forced to be higher than that of the other sample, by a margin of at least $m=0.1$. This unties the uncertainty values from the scale of the task loss, improving on the flexibility principle (iii).\footnote{Being scale-free also means being uncalibrated. However, this is not a disadvantage for pretrained uncertainties, because during pretraining the downstream task is unknown, hence it is impossible to be calibrated for the unseen downstream task in the first place.} 

\section{Experiments} \label{sec:quant}

We now study the main objective of pretrained uncertainties, their performance on downstream datasets. Additionally, we investigate which types of uncertainties they represent.

\subsection{Experimental Setup} 

We are interested in how good our uncertainties perform on unseen datasets. This challenging zero-shot transfer is simple with our pretrained uncertainties, our enhanced loss prediction module outputs an uncertainty $u(x)$ for every image $x$ from the downstream task. In order to measure the quality of these uncertainty estimates, we follow URL \cite{kirchhof2023url} in using the representation AUROC (R-AUROC) metric. It performs a 1-nearest neighbor classification on the representations $e(x)$ on all images of a downstream dataset. The uncertainties $u(x)$ should then be higher for images that are misclassified, which is quantified by the area under the ROC curve between the predicted uncertainties and whether or not the representation is correct in the sense that it is placed next to another representations of the same class. The R-AUROC can benchmark uncertainties when the classes are unseen during training, but if classes are seen during training, it is highly correlated with a conventional classification AUROC \cite{kirchhof2023url}. In all experiments we also report the 1-nearest neighbor accuracy (Recall@1) from representation learning to quantify the retrieval performance of the representations and verify the non-interference principle (i) above.

We focus on Vision Transformers \cite{vit} of several sizes and report results for the ViT-Base unless otherwise noted. Their backbone and classifiers were already pretrained by \cite{steiner2021augreg} on ImageNet-21k-Winter-2021 (ImageNet-21k-W) \cite{deng2009imagenet} in \texttt{timm} \cite{rw2019timm}, so we only train the uncertainty module. As we show in \cref{sec:negative}, our approach is robust to architecture and optimizer hyperparameters, so we use the default values reported in \cref{app:details}, inter alia a lightweight 2-layer MLP with width 512 for the uncertainty head. We train each model on five seeds and report the median as well as the distance to the maximum or minimum, whichever is larger, to provide an interpretable means to judge the variation.

As datasets, we use ImageNet-21k-W for pretraining and twelve datasets that span a variety of natural image domains for zero-shot transfer. Three are used in the URL benchmark, namely CUB-200-2011 \cite{cub200}, CARS196 \cite{cars196}, and Stanford Online Products (SOP) \cite{song2016deep}. Seven are the natural images datasets of the Visual Task Adaption Benchmark (VTAB) \cite{zhai2020largescale}, namely Caltech101 \cite{FeiFei2004LearningGV}, Oxford IIIT Pets \cite{parkhi12a}, CIFAR100 \cite{Krizhevsky09learningmultiple}, Scene Understanding 397 (SUN) \cite{Xiao2010}, Oxford Flowers 102 \cite{Nilsback08}, Describable Textures (DTD) \cite{cimpoi14describing}, and Street View House Numbers (SVHN) \cite{Netzer2011}. The remaining two are CIFAR10 \cite{Krizhevsky09learningmultiple} and Treeversity\#1 \cite{schmarje2022one}. 

\begin{figure*}[t]
    \centering
    \includegraphics{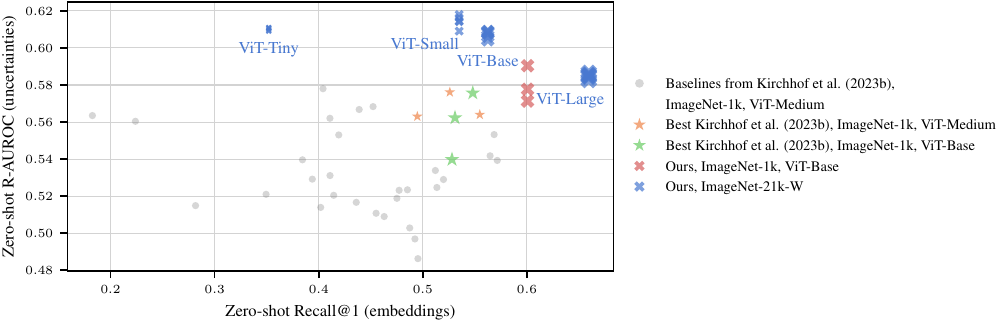}
    \caption{Our pretrained uncertainties outperform the approaches in the URL benchmark \cite{kirchhof2023url}. The URL benchmark trained ViT-Mediums on ImageNet-1k. We reimplement its best approach (orange) on ViT-Base (green), then enhance it with our changes (red), and finally scale the training of ours to ImageNet-21k with various ViT sizes (blue). Each dot is one seed.}
    \label{fig:scaling}
\end{figure*}

\begin{figure*}[t]
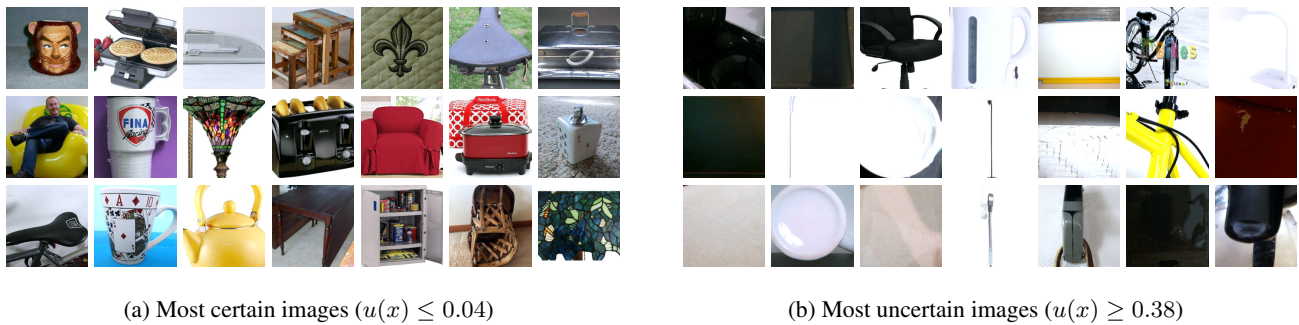

  \centering \vspace{5mm}
  \begin{subfigure}[b]{0.47\linewidth}
  \centering
  \begin{tikzpicture}
    \foreach \row in {0,1,2} {
      \foreach \col in {0,1,2,3,4,5,6} {
        \pgfmathtruncatemacro\imageindex{\row * 7 + \col + 1}
        \node[anchor=south west,inner sep=0] at (\col*1.18,-\row*1.18) {\includegraphics[width=1.08cm,height=1.08cm]{figures/lowunc_\imageindex.jpg}};
      }
    }
  \end{tikzpicture} \vspace{-2mm}
  \caption{\footnotesize Most certain images ($u(x) \leq 0.04$)}
  \label{fig:most_certain}
  \end{subfigure}
  \hfill
  \begin{subfigure}[b]{0.47\linewidth}
  \centering
  \begin{tikzpicture}
    \foreach \row in {0,1,2} {
      \foreach \col in {0,1,2,3,4,5,6} {
        \pgfmathtruncatemacro\imageindex{\row * 7 + \col + 1}
        \node[anchor=south west,inner sep=0] at (\col*1.18,-\row*1.18) {\includegraphics[width=1.08cm,height=1.08cm]{figures/highunc_\imageindex.jpg}};
      }
    }
  \end{tikzpicture} \vspace{-2mm}
  \caption{\footnotesize Most uncertain images ($u(x) \geq 0.38$)}
  \label{fig:most_uncertain}
  \end{subfigure} \vspace{-1mm}
  \caption{Pretrained uncertainties separate clear from ambiguous images on Stanford Online Products, a zero-shot dataset.}
  \label{fig:examples}
\end{figure*}

\subsection{Pretrained Uncertainties Generalize} We first test the generalization principle (ii) on the twelve unseen datasets. \cref{fig:generalize} shows that our pretrained uncertainties generalize well on eleven of the twelve datasets. The best R-AUROCs (Caltech101: \res{0.758}{0.006}, Oxford Pets: \res{0.740}{0.008}, and CIFAR10: \res{0.739}{0.002}) are close to that of the pretraining dataset (\res{0.791}{0.001}). In \cref{app:generalize} we find that the zero-shot R-AUROC is higher on datasets that are closer to the domain spanned by ImageNet-21k-W, as one would expect from a pretrained model. This implies that further scaling the pretraining corpus, which is possible with our efficient training, may further benefit performance. The performance also depends on the granulatity of the zero-shot dataset. SVHN for example demands fine-grained house number disambiguation. It is harder to assign a pretrained uncertainty to such specialized tasks without knowing them in advance.

\subsection{A New State-of-the-art} 
How do these results compare to the transfer performances of other methods in the field? The URL benchmark \cite{kirchhof2023url} has recently tested the R-AUROC of eleven approaches on the CUB, CARS, and SOP datasets, averaged. URL tested the methods on ViT-Medium, a niche architecture that does not have ImageNet-21k checkpoints available. Thus, we switch to ViT-Base and reimplement URL's best performing method, the original loss prediction from \cref{sec:losspred}. We reuse their codebase for compatibility. 

\cref{fig:scaling} shows the results, both in terms of R-AUROC and Recall@1. First, we find that our ViT-Base loss prediction reimplementation (red stars) achieves comparable performance to the original ViT-Medium backbone (orange stars). We then enhance the original loss prediction with our changes from \cref{sec:enhance} (green crosses). We find that the Recall@1 increases by \res{0.065}{0.012} because the backbone is no longer deteriorated by the uncertainty module gradients. The Recall@1 is now constant because we only train the uncertainty head anymore, keeping the pretrained backbone frozen. This does not restrict the R-AUROC of the uncertainty head, it in fact increases slightly by \res{0.021}{0.030}. Finally, we scale from ImageNet-1k to ImageNet-21k-W pretraining (blue crosses). This increases the R-AUROC again by \res{0.028}{0.014} on the ViT-Base. Because we now use a different ImageNet-21k-W pretrained checkpoint, its Recall@1 changes. Upon training uncertainty modules for various ViT sizes, we find that they all have higher R-AUROC than the previous state-of-the-art. This demonstrates the generality of our approach.

\begin{figure*}[t]
    \centering
    \begin{tikzpicture}
    \node[anchor=south west,inner sep=0] (image) at (0,0) {\includegraphics{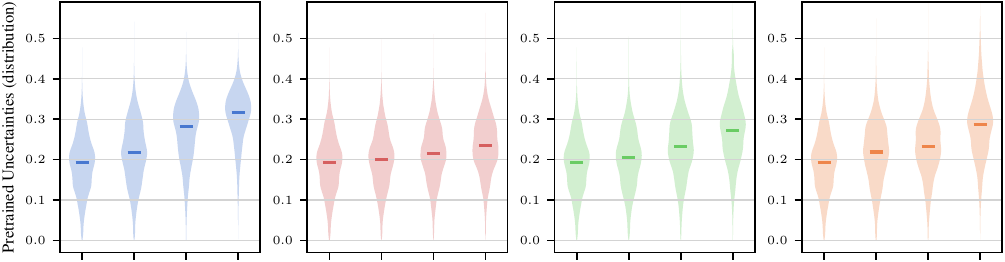}};
    \begin{scope}[x={(image.south east)},y={(image.north west)}]
        \node[anchor=north,inner sep=0] (image) at (0.082,0) {\includegraphics[scale=0.07]{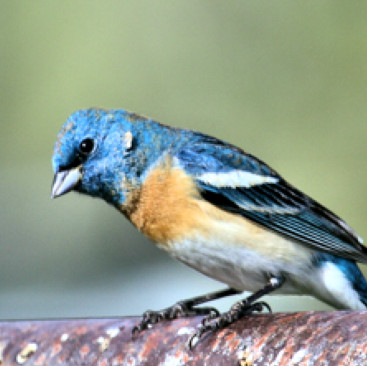}};
        \node[anchor=north,inner sep=0] (image) at (0.133,0) {\includegraphics[scale=0.07]{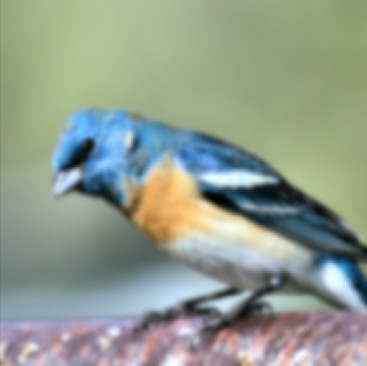}};
        \node[anchor=north,inner sep=0] (image) at (0.186,0) {\includegraphics[scale=0.07]{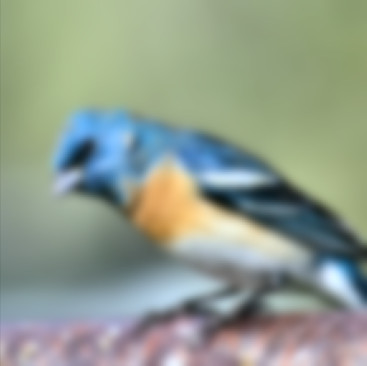}};
        \node[anchor=north,inner sep=0] (image) at (0.238,0) {\includegraphics[scale=0.07]{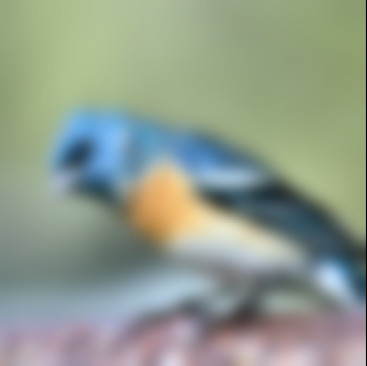}};
        \node[anchor=north,inner sep=0] (image) at (0.329,0) {\includegraphics[scale=0.07]{figures/example_baseline.jpg}};
        \node[anchor=north,inner sep=0] (image) at (0.380,0) {\includegraphics[scale=0.07]{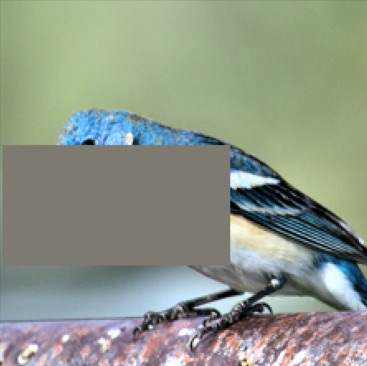}};
        \node[anchor=north,inner sep=0] (image) at (0.432,0) {\includegraphics[scale=0.07]{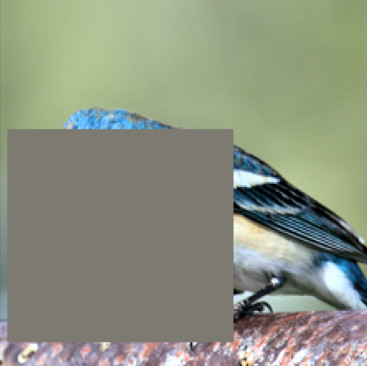}};
        \node[anchor=north,inner sep=0] (image) at (0.483,0) {\includegraphics[scale=0.07]{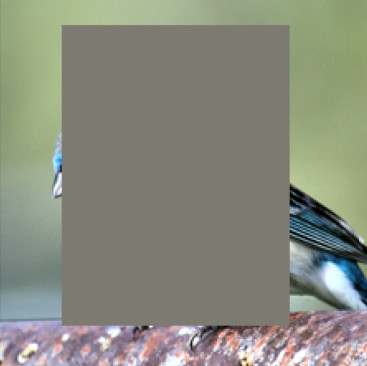}};
        \node[anchor=north,inner sep=0] (image) at (0.574,0) {\includegraphics[scale=0.07]{figures/example_baseline.jpg}};
        \node[anchor=north,inner sep=0] (image) at (0.627,0) {\includegraphics[scale=0.07]{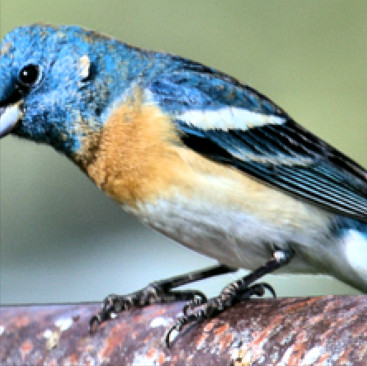}};
        \node[anchor=north,inner sep=0] (image) at (0.679,0) {\includegraphics[scale=0.07]{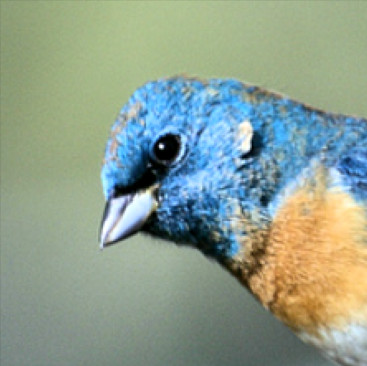}};
        \node[anchor=north,inner sep=0] (image) at (0.730,0) {\includegraphics[scale=0.07]{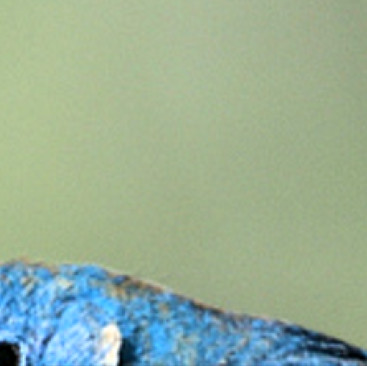}};
        \node[anchor=north,inner sep=0] (image) at (0.821,0) {\includegraphics[scale=0.07]{figures/example_baseline.jpg}};
        \node[anchor=north,inner sep=0] (image) at (0.872,0) {\includegraphics[scale=0.07]{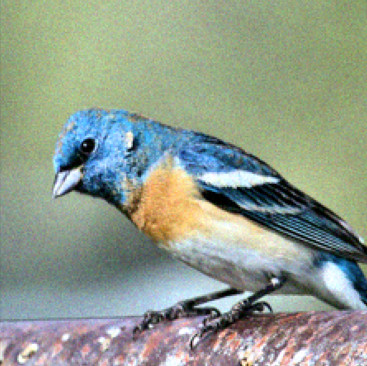}};
        \node[anchor=north,inner sep=0] (image) at (0.924,0) {\includegraphics[scale=0.07]{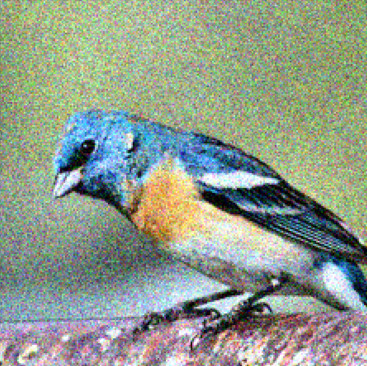}};
        \node[anchor=north,inner sep=0] (image) at (0.977,0) {\includegraphics[scale=0.07]{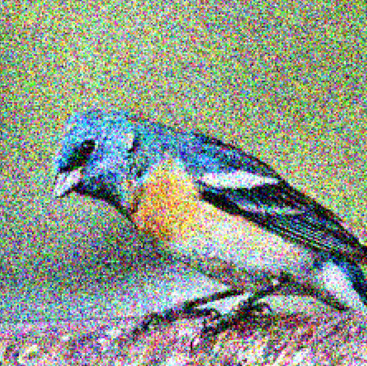}};
        \node [anchor=north] at (0.16,-0.19) {\footnotesize (a) Blur};
        \node [anchor=north] at (0.407,-0.19) {\footnotesize (b) Boxes};
        \node [anchor=north] at (0.655,-0.19) {\footnotesize (c) Zoom};
        \node [anchor=north] at (0.898,-0.19) {\footnotesize (d) Noise};
    \end{scope}
    \end{tikzpicture}
    \vspace{-2mm}
    \caption{Pretrained uncertainties grow as images are deteriorated. Distributions and medians over unseen datasets (CUB, CARS, SOP). Note that except zooming, the pretrained model is not exposed to these data augmentations during training.}
    \label{fig:deteriorate}
\end{figure*}

\begin{figure}[t]
    \centering \vspace{3mm}
    \includegraphics{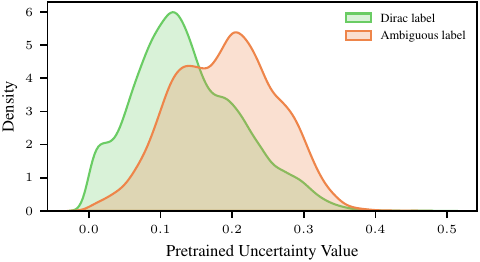}
    \caption{Pretrained uncertainties are systematically higher for ImageNet ReaL-H images with multiple possible labels.
    }
    \label{fig:ambiguous}
\end{figure}

\begin{figure}[t]
    \centering \vspace{3mm}
    \includegraphics{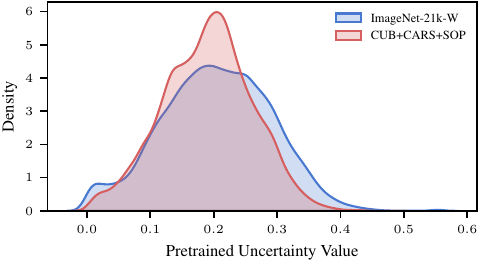}
    \caption{Pretrained uncertainties are consistent between train and unseen datasets, indicating the absence of epistemic uncertainty.}
    \label{fig:idood}
\end{figure}

\subsection{Negative Results: Simple Beats Complex} \label{sec:negative}
Before we continue, we share some negative results. In \cref{app:negative}, we experiment with several techniques to further improve our method, including softening the loss function, uncertainty-induced training data augmentations, architecture changes, optimizer modifications, and initialization schemes. However, none of them significantly improve the uncertainties beyond the method presented in \cref{sec:enhance}. Thus, to avoid adding unnecessary complexity, we decide to keep our approach as clean and simple as it is.

\subsection{Pretrained Uncertainties $\approx$ Aleatoric Uncertainty} \label{sec:understand}

If pretrained uncertainties work on unseen datasets, then which uncertainties do they capture? In this section, we find that they model primarily aleatoric uncertainty and are mostly invariant to epistemic uncertainty. We conduct all analyses on unseen datasets and the ViT-Base model, unless otherwise noted.

To form a working hypothesis, we give some randomly selected examples with low and high predicted uncertainty in \cref{fig:examples}. Although the network was not trained on this task, eBay product images, it correctly gives low uncertainties to clear and high uncertainties to ambiguous images.
Not even a human expert or a Bayes classifier will be able to reduce the ambiguous images' uncertainty to zero, their ambiguity is intrinsic.
This is known as aleatoric uncertainty. We hypothesize that pretrained uncertainties represent this form of uncertainty. Below, we investigate this hypothesis.

\textbf{Human ambiguities.} Aleatoric uncertainty is what is left even when an expert makes a prediction, for example a human annotator. So, we compare the model uncertainties to those of human annotators. ImageNet-1k ReaL-H \cite{beyer2020we} re-collected labels for the 50,000 images in the ImageNet-1k validation set.\footnote{Our pretraining dataset ImageNet-21k-W covers the classes of ImageNet-1k but neither its validation images nor any soft labels.} While clear images kept their original Dirac label, annotators gave multiple or no labels to an image if it was ambiguous. \cref{fig:ambiguous} shows that pretrained uncertainties are systematically higher on images the human annotators considered ambiguous (AUROC 0.701). This reinforces the aleatoric uncertainty hypothesis.

\textbf{Interventional study.} Second, we run an interventional experiment to induce aleatoric uncertainty. We deteriorate the images of the unseen datasets by blurring, overlaying with grey boxes, zooming in strongly, and adding Gaussian noise. Except zooming, these transformations were not applied during pretraining. \cref{fig:deteriorate} shows that each of these transformations increases the pretrained uncertainties, the more strongly we deteriorate the images. This is additional evidence for the aleatoric uncertainty hypothesis.

\begin{figure*}[t]
    \centering \vspace{-3mm}
    \begin{tikzpicture}
    \node[anchor=south west,inner sep=0] (image) at (0,0) {\includegraphics[width=0.795\linewidth]{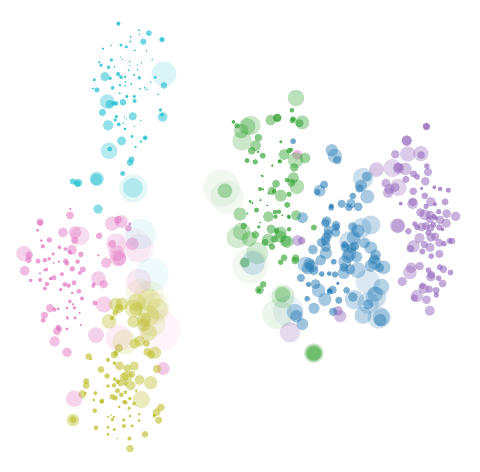}};
    \begin{scope}[x={(image.south east)},y={(image.north west)}]
        \node[anchor=north west,inner sep=0] (image1) at (-0.07,0.71) {\includegraphics[scale=0.25]{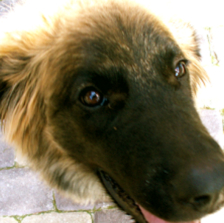}};
        \node[below=1mm of image1,inner sep=0] (text1) {\footnotesize$u(x)=0.243$};
        \draw[blue] (image1.south east) -- (0.293, 0.44);
        \node[anchor=north west,inner sep=0] (image2) at (-0.07,0.89) {\includegraphics[scale=0.25]{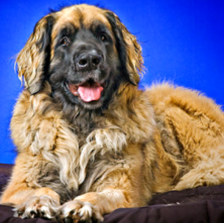}};
        \node[below=1mm of image2,inner sep=0] (text2) {\footnotesize$u(x)=0.046$};
        \draw[blue] (image2.south east) -- (0.263, 0.797);
        \node[anchor=north west,inner sep=0] (image3) at (0.7,0.95) {\includegraphics[scale=0.25]{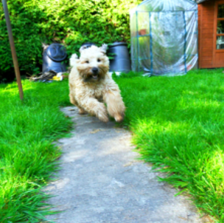}};
        \node[below=1mm of image3,inner sep=0] (text3) {\footnotesize$u(x)=0.147$};
        \draw[green] (image3.south west) -- (0.47, 0.61);
        \node[anchor=north west,inner sep=0] (image4) at (0.88,0.95) {\includegraphics[scale=0.25]{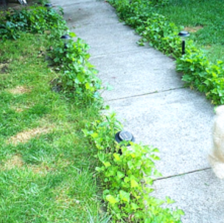}};
        \node[below=1mm of image4,inner sep=0] (text4) {\footnotesize$u(x)=0.256$};
        \draw[green] (image4.south west) -- (0.482, 0.63);
        \node[anchor=north west,inner sep=0] (image5) at (1.05,0.73) {\includegraphics[scale=0.25]{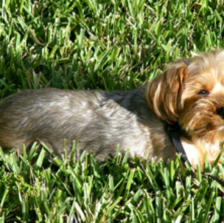}};
        \node[below=1mm of image5,inner sep=0] (text5) {\footnotesize$u(x)=0.169$};
        \draw[purple] (image5.south west) -- (0.813, 0.626);
        \node[anchor=north west,inner sep=0] (image6) at (1.05,0.55) {\includegraphics[scale=0.25]{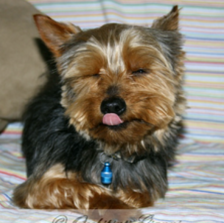}};
        \node[below=1mm of image6,inner sep=0] (text6) {\footnotesize $u(x)=0.051$};
        \draw[purple] (image6.south west) -- (0.922, 0.515);
    \end{scope}
    \end{tikzpicture} \vspace{-4mm}
    \caption{Visualizing pretrained uncertainties makes it easy to identify outliers in a tSNE plot. Images with high uncertainty $u(x)$ are larger and transparent. Six classes of the zero-shot Oxford Pets dataset.}
    \label{fig:clustering}
\end{figure*}
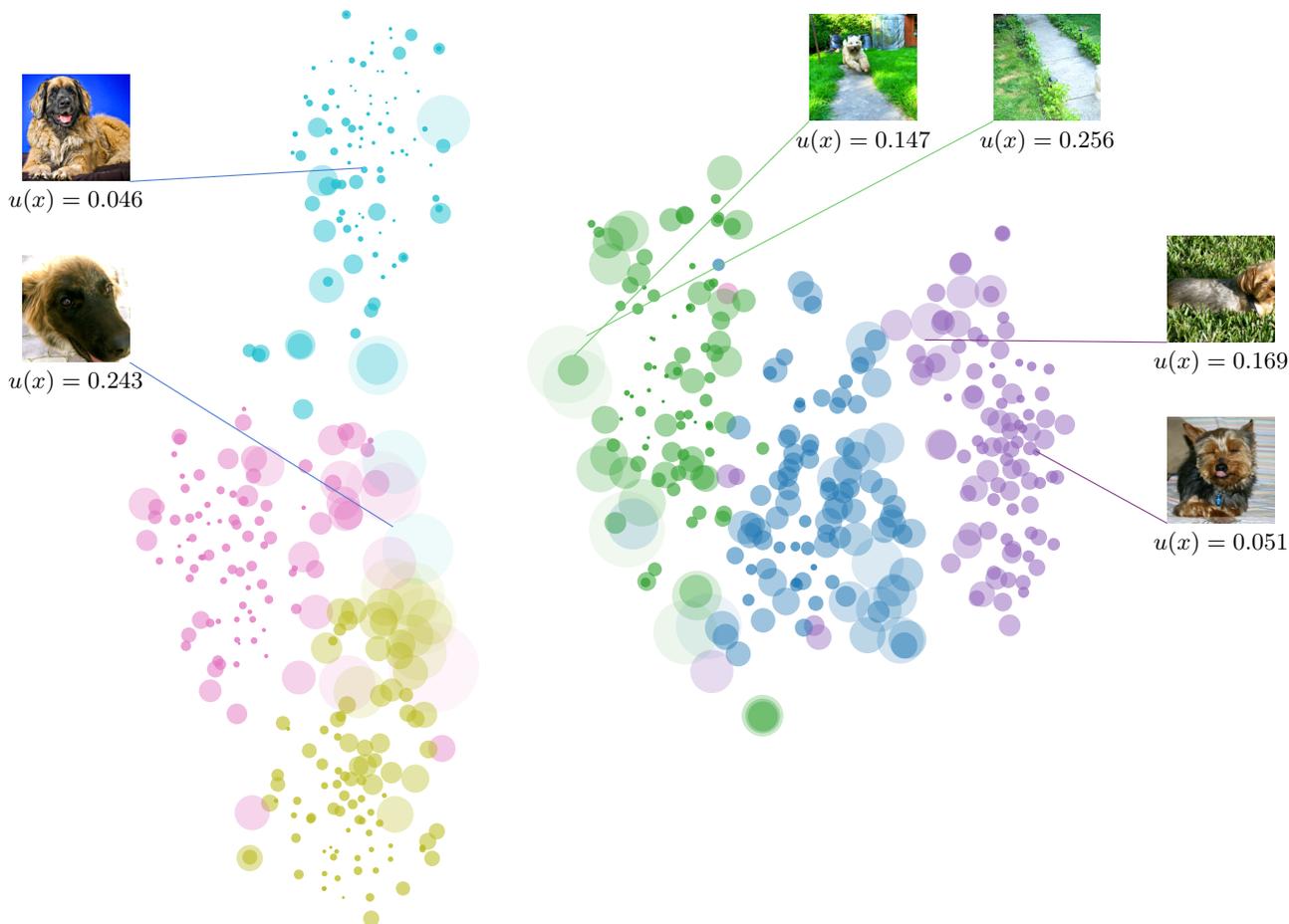

\textbf{No sign of epistemic uncertainty.} We now consider the opposite hypothesis: Besides aleatoric uncertainty, do pretrained uncertainties comprise epistemic uncertainty?
This uncertainty arises when a model has not seen an input before. Note that this would be detrimental to a pretrained uncertainty model, since it is intended to be used (exclusively) on unseen datasets where every image would be highly epistemically uncertain, drowning out the aleatoric signal. 
To test for epistemic uncertainty, we compare the pretrained uncertainties of in-distribution pretraining images to those of out-of-distribution images from unseen datasets. \cref{fig:idood} shows that the uncertainties on the pretraining data are similarly distributed to the unseen dataset (pairwise AUROC \res{0.503}{0.004}). This suggests they are primarily capturing aleatoric uncertainty.

In summary, our results suggest that pretrained uncertainties quantify the amount of aleatoric uncertainty in an image, both in- and out-of-distribution, without being confounded by epistemic uncertainty. This constitutes significant progress for the ongoing efforts to disentangle epistemic from aleatoric uncertainty \cite{wimmer2023quantifying,valdenegro2022deeper}.

\section{Application Examples} \label{sec:applications}

In this section we showcase two applications that are unlocked by our new pretrained uncertainties.

\subsection{Uncertainty-aware tSNE}

Pretrained representations are often used to visualize datasets using methods like tSNE \cite{JMLR:v9:vandermaaten08a} or UMAP \cite{McInnes2018}. With pretrained uncertainties, we can now communicate inhererent ambiguities and explain outliers in these plots.

\cref{fig:clustering} shows a tSNE visualization of six dog breeds in Oxford Pets, an unseen dataset. If an image has a high pretrained uncertainty, it is plotted as a larger and increasingly transparent circle. The core region of each class cluster, typified by many small opaque circles, is now visually distinct from border regions and outliers, which are typified by collections of large transparent circles. By inspecting the images that underlie each representation, we verify that the core regions with low uncertainties comprise prototypical images whereas more uncertain images are often cropped out or in a camera angle that makes the exact dog breed ambiguous. We can also see that images lying in other classes' regions are often highly uncertain. Such misclassifications can be prevented by using our uncertainty-enhanced tSNE plots by allowing practitioners to understand and adjust the data preprocessing and filtering. 

\subsection{Safe Retrieval}

This outlier identification can also be automated and utilized to make image retrieval more robust, enabling safe retrieval.

Consider again the Oxford Pets dataset in \cref{fig:clustering}. If we add a new dog image and search for its nearest neighbor, existing next-neighbor retrieval systems \cite{douze2024faiss} may match it to an ambiguous image since these tend to lay at border regions. Similarly, if our new image itself is ambiguous, it is likely misplaced and existing systems will match it an arbitrary class. We can utilize pretrained uncertainties to tackle both of these problems by 
\begin{enumerate}
    \item Rejecting queries that are uncertain, and
    \item Removing ambiguous images in the existing dataset, making it impossible to match to them.
\end{enumerate}
As an example, we reject and/or clean the $10\%$ most uncertain images per class in Oxford Pets. \cref{tab:saferetrieval} shows that a typical cosine-distance based next-neighbour search achieves a Recall@1 of \res{0.772}{0.000}, or in other words a rate of \res{0.228}{0.000} wrong retrievals. Refusing to retrieve images when the input query is uncertain reduces this error rate by 14\% to \res{0.196}{0.003} and cleaning the dataset from ambiguous images as potential retrieval partners reduces it by an additional 17\%. These improvements are not only observed on unseen dataset but also on data that the retrieval system is familiar with. When using the ImageNet-1k validation set, whose classes were seen during ImageNet-21k-W pretraining (but whose validation images are unseen), deferring ambiguous queries reduces the error rate by 10\% and cleaning the database reduces another 10\%. All of these improvements are obtained automatically and fully unsupervised - we do not need to know the ground truth label of either the input or the database, since pretrained uncertainties can be computed for any input.

These are only first demonstrations of the opportunities that pretrained uncertainties offer. We anticipate further applications building up on pretrained uncertainties. For example, recent literature proposes retrieving a set of potential neighbours that is close to an ambiguous input with respect to its representation uncertainty \cite{kirchhof2023probabilistic}. This can be implemented with pretrained uncertainty since they give uncertainties about representations. Similarly, conformal prediction \cite{angelopoulos2022gentle} can view our pretrained uncertainties as a scoring function and calibrate its uncertainty predictions to downstream datasets. To facilitate future research, we provide all pretrained uncertainty checkpoints and code under the link in the abstract.

\begin{table}[t]
    \centering
    \footnotesize
    \begin{tabular}{lcc}
    \toprule
         1-NN error & Oxford Pets & ImageNet-1k \\
    \cmidrule(lr){1-3}
         Full datasets & \res{0.228}{0.000} & \res{0.382}{0.000} \\
         + clean queries & \res{0.196}{0.003} & \res{0.343}{0.001} \\
         + clean database & \res{0.163}{0.003} &\res{0.307}{0.001}\\
    \bottomrule
    \end{tabular}
    \caption{Pretrained uncertainties reduce the error rate of next-neighbour retrieval systems by rejecting ambiguous queries and/or removing ambiguous images from the database.}
    \label{tab:saferetrieval}
    
\end{table}

\section{Conclusion}
This work introduces a pretrained uncertainty module for computer vision models that is simple, cheap and scalable. We demonstrate its scalability by pretraining on ImageNet-21k-W. Our pretrained uncertainties method gives state-of-the-art zero shot uncertainty estimates on unseen datasets i.e.\ without finetuning. In future work, we anticipate scaling to even larger pretraining datasets as well as extending the method to pretraining objectives beyond classification and beyond the vision domain. We expect our fixes to the vanilla loss prediction method, that eliminate interference between uncertainty prediction and the main task, to also help in other feed-forward uncertainty quantifiers. By providing pretrained checkpoints, we intend to support applications similar to enhanced visualization and safe retrieval.

\section*{Impact Statement}
Our uncertainties are intended to capture errors before they happen and reveal uncertainties that would remain undetected when images are solely expressed as representation vectors. This makes models more safe and trustworthy by allowing them to fulfill their tasks with less errors. We enable an easier access to these uncertainties by our ease-of-use principles and providing plug-and-play checkpoints. We see this as a positive impact on both the community and society. As with all general-purpose machine learning advancements, this assumes that a practitioner does not develop a model with a harmful task, which is beyond our sphere of influence. Additionally, we encourage researchers to follow our example of finding ways to significantly reduce training costs for a lower energy consumption during training. We provide our code standalone to help start these efforts.

\bibliography{main}

\begin{thebibliography}{51}
\providecommand{\natexlab}[1]{#1}
\providecommand{\url}[1]{\texttt{#1}}
\expandafter\ifx\csname urlstyle\endcsname\relax
  \providecommand{\doi}[1]{doi: #1}\else
  \providecommand{\doi}{doi: \begingroup \urlstyle{rm}\Url}\fi

\bibitem[Angelopoulos \& Bates(2022)Angelopoulos and Bates]{angelopoulos2022gentle}
Angelopoulos, A.~N. and Bates, S.
\newblock A gentle introduction to conformal prediction and distribution-free uncertainty quantification.
\newblock \emph{arXiv preprint arXiv:2107.07511}, 2022.

\bibitem[Beyer et~al.(2020)Beyer, H{\'e}naff, Kolesnikov, Zhai, and Oord]{beyer2020we}
Beyer, L., H{\'e}naff, O.~J., Kolesnikov, A., Zhai, X., and Oord, A. v.~d.
\newblock Are we done with {ImageNet}?
\newblock \emph{arXiv preprint arXiv:2006.07159}, 2020.

\bibitem[Chen et~al.(2023)Chen, Liang, Huang, Real, Wang, Liu, Pham, Dong, Luong, Hsieh, et~al.]{chen2023symbolic}
Chen, X., Liang, C., Huang, D., Real, E., Wang, K., Liu, Y., Pham, H., Dong, X., Luong, T., Hsieh, C.-J., et~al.
\newblock Symbolic discovery of optimization algorithms.
\newblock \emph{arXiv preprint arXiv:2302.06675}, 2023.

\bibitem[Chun(2023)]{chun2023improved}
Chun, S.
\newblock Improved probabilistic image-text representations.
\newblock \emph{arXiv preprint arXiv:2305.18171}, 2023.

\bibitem[Cimpoi et~al.(2014)Cimpoi, Maji, Kokkinos, Mohamed, and Vedaldi]{cimpoi14describing}
Cimpoi, M., Maji, S., Kokkinos, I., Mohamed, S., and Vedaldi, A.
\newblock Describing textures in the wild.
\newblock In \emph{Computer Vision and Pattern Recognition ({CVPR})}, 2014.

\bibitem[Collier et~al.(2023)Collier, Jenatton, Mustafa, Houlsby, Berent, and Kokiopoulou]{collier2023massively}
Collier, M., Jenatton, R., Mustafa, B., Houlsby, N., Berent, J., and Kokiopoulou, E.
\newblock Massively scaling heteroscedastic classifiers.
\newblock \emph{arXiv preprint arXiv:2301.12860}, 2023.

\bibitem[Cui et~al.(2023)Cui, Zhang, Deng, Dong, and Zhu]{cui2023learning}
Cui, P., Zhang, D., Deng, Z., Dong, Y., and Zhu, J.
\newblock Learning sample difficulty from pre-trained models for reliable prediction.
\newblock In \emph{Neural Information Processing Systems (NeurIPS)}, 2023.

\bibitem[Dehghani et~al.(2023)Dehghani, Djolonga, Mustafa, Padlewski, Heek, Gilmer, Steiner, Caron, Geirhos, Alabdulmohsin, et~al.]{dehghani2023scaling}
Dehghani, M., Djolonga, J., Mustafa, B., Padlewski, P., Heek, J., Gilmer, J., Steiner, A.~P., Caron, M., Geirhos, R., Alabdulmohsin, I., et~al.
\newblock Scaling vision transformers to 22 billion parameters.
\newblock In \emph{International Conference on Machine Learning (ICML)}, 2023.

\bibitem[Deng et~al.(2009)Deng, Dong, Socher, Li, Li, and Fei-Fei]{deng2009imagenet}
Deng, J., Dong, W., Socher, R., Li, L.-J., Li, K., and Fei-Fei, L.
\newblock {ImageNet}: A large-scale hierarchical image database.
\newblock In \emph{Computer Vision and Pattern Recognition (CVPR)}, 2009.

\bibitem[Dosovitskiy et~al.(2021)Dosovitskiy, Beyer, Kolesnikov, Weissenborn, Zhai, Unterthiner, Dehghani, Minderer, Heigold, Gelly, Uszkoreit, and Houlsby]{vit}
Dosovitskiy, A., Beyer, L., Kolesnikov, A., Weissenborn, D., Zhai, X., Unterthiner, T., Dehghani, M., Minderer, M., Heigold, G., Gelly, S., Uszkoreit, J., and Houlsby, N.
\newblock An image is worth 16x16 words: Transformers for image recognition at scale.
\newblock In \emph{International Conference on Learning Representations (ICLR)}, 2021.

\bibitem[Douze et~al.(2024)Douze, Guzhva, Deng, Johnson, Szilvasy, Mazaré, Lomeli, Hosseini, and Jégou]{douze2024faiss}
Douze, M., Guzhva, A., Deng, C., Johnson, J., Szilvasy, G., Mazaré, P.-E., Lomeli, M., Hosseini, L., and Jégou, H.
\newblock The {F}aiss library.
\newblock \emph{arXiv preprint arXiv:2401.08281}, 2024.

\bibitem[Fei-Fei et~al.(2004)Fei-Fei, Fergus, and Perona]{FeiFei2004LearningGV}
Fei-Fei, L., Fergus, R., and Perona, P.
\newblock Learning generative visual models from few training examples: An incremental {B}ayesian approach tested on 101 object categories.
\newblock \emph{Computer Vision and Pattern Recognition Workshop}, 2004.

\bibitem[Galil et~al.(2023{\natexlab{a}})Galil, Dabbah, and El-Yaniv]{galil2023a}
Galil, I., Dabbah, M., and El-Yaniv, R.
\newblock A framework for benchmarking class-out-of-distribution detection and its application to {ImageNet}.
\newblock In \emph{The Eleventh International Conference on Learning Representations (ICLR)}, 2023{\natexlab{a}}.

\bibitem[Galil et~al.(2023{\natexlab{b}})Galil, Dabbah, and El-Yaniv]{galil2023what}
Galil, I., Dabbah, M., and El-Yaniv, R.
\newblock What can we learn from the selective prediction and uncertainty estimation performance of 523 {ImageNet} classifiers?
\newblock In \emph{International Conference on Learning Representations (ICLR)}, 2023{\natexlab{b}}.

\bibitem[Hendrycks et~al.(2020)Hendrycks, Mu, Cubuk, Zoph, Gilmer, and Lakshminarayanan]{hendrycks2020augmix}
Hendrycks, D., Mu, N., Cubuk, E.~D., Zoph, B., Gilmer, J., and Lakshminarayanan, B.
\newblock {AugMix}: A simple method to improve robustness and uncertainty under data shift.
\newblock In \emph{International Conference on Learning Representations (ICLR)}, 2020.

\bibitem[Karpukhin et~al.(2022)Karpukhin, Dereka, and Kolesnikov]{karpukhin2022probabilistic}
Karpukhin, I., Dereka, S., and Kolesnikov, S.
\newblock Probabilistic embeddings revisited.
\newblock \emph{arXiv preprint arXiv:2202.06768}, 2022.

\bibitem[Kim et~al.()Kim, Jung, Park, Kim, and Yoon]{pmlr-v202-kim23g}
Kim, E., Jung, D., Park, S., Kim, S., and Yoon, S.
\newblock Probabilistic concept bottleneck models.
\newblock In \emph{International Conference on Machine Learning (ICML)}.

\bibitem[Kingma \& Ba(2015)Kingma and Ba]{kingma2014adam}
Kingma, D.~P. and Ba, J.
\newblock Adam: {A} method for stochastic optimization.
\newblock \emph{International Conference on Learning Representations (ICLR)}, 2015.

\bibitem[Kirchhof et~al.(2022)Kirchhof, Roth, Akata, and Kasneci]{kirchhof2022non}
Kirchhof, M., Roth, K., Akata, Z., and Kasneci, E.
\newblock A non-isotropic probabilistic take on proxy-based deep metric learning.
\newblock In \emph{European Conference on Computer Vision (ECCV)}, 2022.

\bibitem[Kirchhof et~al.(2023{\natexlab{a}})Kirchhof, Kasneci, and Oh]{kirchhof2023probabilistic}
Kirchhof, M., Kasneci, E., and Oh, S.~J.
\newblock Probabilistic contrastive learning recovers the correct aleatoric uncertainty of ambiguous inputs.
\newblock \emph{International Conference on Machine Learning (ICML)}, 2023{\natexlab{a}}.

\bibitem[Kirchhof et~al.(2023{\natexlab{b}})Kirchhof, Mucsányi, Oh, and Kasneci]{kirchhof2023url}
Kirchhof, M., Mucsányi, B., Oh, S.~J., and Kasneci, E.
\newblock Url: A representation learning benchmark for transferable uncertainty estimates.
\newblock \emph{Proceedings of the Neural Information Processing Systems Track on Datasets and Benchmarks}, 2023{\natexlab{b}}.

\bibitem[Krause et~al.(2013)Krause, Stark, Deng, and Fei-Fei]{cars196}
Krause, J., Stark, M., Deng, J., and Fei-Fei, L.
\newblock {3D} object representations for fine-grained categorization.
\newblock In \emph{Conference on Computer Vision and Pattern Recognition (CVPR) Workshop}, 2013.

\bibitem[Krizhevsky(2009)]{Krizhevsky09learningmultiple}
Krizhevsky, A.
\newblock Learning multiple layers of features from tiny images.
\newblock Technical report, 2009.

\bibitem[Lahlou et~al.(2023)Lahlou, Jain, Nekoei, Butoi, Bertin, Rector-Brooks, Korablyov, and Bengio]{lahlou2023deup}
Lahlou, S., Jain, M., Nekoei, H., Butoi, V.~I., Bertin, P., Rector-Brooks, J., Korablyov, M., and Bengio, Y.
\newblock {DEUP}: Direct epistemic uncertainty prediction.
\newblock \emph{Transactions on Machine Learning Research (TMLR)}, 2023.
\newblock ISSN 2835-8856.

\bibitem[Laves et~al.(2020)Laves, Ihler, Fast, Kahrs, and Ortmaier]{laves2020well}
Laves, M.-H., Ihler, S., Fast, J.~F., Kahrs, L.~A., and Ortmaier, T.
\newblock Well-calibrated regression uncertainty in medical imaging with deep learning.
\newblock In \emph{Medical Imaging with Deep Learning}, pp.\  393--412. PMLR, 2020.

\bibitem[Liu et~al.(2020)Liu, Lin, Padhy, Tran, Bedrax~Weiss, and Lakshminarayanan]{liu2020simple}
Liu, J., Lin, Z., Padhy, S., Tran, D., Bedrax~Weiss, T., and Lakshminarayanan, B.
\newblock Simple and principled uncertainty estimation with deterministic deep learning via distance awareness.
\newblock \emph{Advances in Neural Information Processing Systems (NeurIPS)}, 2020.

\bibitem[Loshchilov \& Hutter(2017)Loshchilov and Hutter]{loshchilov2017decoupled}
Loshchilov, I. and Hutter, F.
\newblock Decoupled weight decay regularization.
\newblock \emph{arXiv preprint arXiv:1711.05101}, 2017.

\bibitem[McInnes et~al.(2018)McInnes, Healy, Saul, and Großberger]{McInnes2018}
McInnes, L., Healy, J., Saul, N., and Großberger, L.
\newblock {UMAP}: Uniform manifold approximation and projection.
\newblock \emph{Journal of Open Source Software}, 3\penalty0 (29):\penalty0 861, 2018.
\newblock \doi{10.21105/joss.00861}.
\newblock URL \url{https://doi.org/10.21105/joss.00861}.

\bibitem[Mucsányi et~al.(2023)Mucsányi, Kirchhof, Nguyen, Rubinstein, and Oh]{mucsanyi2023trustworthy}
Mucsányi, B., Kirchhof, M., Nguyen, E., Rubinstein, A., and Oh, S.~J.
\newblock Trustworthy machine learning, 2023.

\bibitem[Nakamura et~al.(2023)Nakamura, Okada, and Taniguchi]{nakamura2023representation}
Nakamura, H., Okada, M., and Taniguchi, T.
\newblock Representation uncertainty in self-supervised learning as variational inference.
\newblock In \emph{International Conference on Computer Vision (ICCV)}, 2023.

\bibitem[Netzer et~al.(2011)Netzer, Wang, Coates, Bissacco, Wu, and Ng]{Netzer2011}
Netzer, Y., Wang, T., Coates, A., Bissacco, A., Wu, B., and Ng, A.~Y.
\newblock Reading digits in natural images with unsupervised feature learning.
\newblock 2011.

\bibitem[Nilsback \& Zisserman(2008)Nilsback and Zisserman]{Nilsback08}
Nilsback, M.-E. and Zisserman, A.
\newblock Automated flower classification over a large number of classes.
\newblock In \emph{Proceedings of the Indian Conference on Computer Vision, Graphics and Image Processing}, 2008.

\bibitem[Oh et~al.(2019)Oh, Gallagher, Murphy, Schroff, Pan, and Roth]{oh2018modeling}
Oh, S.~J., Gallagher, A.~C., Murphy, K.~P., Schroff, F., Pan, J., and Roth, J.
\newblock Modeling uncertainty with hedged instance embeddings.
\newblock In \emph{International Conference on Learning Representations (ICLR)}, 2019.

\bibitem[Ovadia et~al.(2019)Ovadia, Fertig, Ren, Nado, Sculley, Nowozin, Dillon, Lakshminarayanan, and Snoek]{ovadia2019can}
Ovadia, Y., Fertig, E., Ren, J., Nado, Z., Sculley, D., Nowozin, S., Dillon, J., Lakshminarayanan, B., and Snoek, J.
\newblock Can you trust your model's uncertainty? {E}valuating predictive uncertainty under dataset shift.
\newblock \emph{Advances in Neural Information Processing Systems (NeurIPS)}, 32, 2019.

\bibitem[Parkhi et~al.(2012)Parkhi, Vedaldi, Zisserman, and Jawahar]{parkhi12a}
Parkhi, O.~M., Vedaldi, A., Zisserman, A., and Jawahar, C.~V.
\newblock Cats and dogs.
\newblock In \emph{Computer Vision and Pattern Recognition (CVPR)}, 2012.

\bibitem[Postels et~al.(2022)Postels, Seg{\`u}, Sun, Sieber, Van~Gool, Yu, and Tombari]{pmlr-v162-postels22a}
Postels, J., Seg{\`u}, M., Sun, T., Sieber, L.~D., Van~Gool, L., Yu, F., and Tombari, F.
\newblock On the practicality of deterministic epistemic uncertainty.
\newblock In \emph{International Conference on Machine Learning (ICML)}, 2022.

\bibitem[Schmarje et~al.(2022)Schmarje, Grossmann, Zelenka, Dippel, Kiko, Oszust, Pastell, Stracke, Valros, Volkmann, et~al.]{schmarje2022one}
Schmarje, L., Grossmann, V., Zelenka, C., Dippel, S., Kiko, R., Oszust, M., Pastell, M., Stracke, J., Valros, A., Volkmann, N., et~al.
\newblock Is one annotation enough? {A} data-centric image classification benchmark for noisy and ambiguous label estimation.
\newblock \emph{arXiv preprint arXiv:2207.06214}, 2022.

\bibitem[Song et~al.(2016)Song, Xiang, Jegelka, and Savarese]{song2016deep}
Song, H.~O., Xiang, Y., Jegelka, S., and Savarese, S.
\newblock Deep metric learning via lifted structured feature embedding.
\newblock In \emph{Computer Vision and Pattern Recognition (CVPR)}, 2016.

\bibitem[Steiner et~al.(2021)Steiner, Kolesnikov, , Zhai, Wightman, Uszkoreit, and Beyer]{steiner2021augreg}
Steiner, A., Kolesnikov, A., , Zhai, X., Wightman, R., Uszkoreit, J., and Beyer, L.
\newblock How to train your {ViT}? {D}ata, augmentation, and regularization in vision transformers.
\newblock \emph{arXiv preprint arXiv:2106.10270}, 2021.

\bibitem[{TorchVision}(2016)]{torchvision2016}
{TorchVision}.
\newblock Torchvision: Pytorch's computer vision library.
\newblock \emph{GitHub repository: \url{https://github.com/pytorch/vision}}, 2016.

\bibitem[Tran et~al.(2022)Tran, Liu, Dusenberry, Phan, Collier, Ren, Han, Wang, Mariet, Hu, et~al.]{tran2022plex}
Tran, D., Liu, J., Dusenberry, M.~W., Phan, D., Collier, M., Ren, J., Han, K., Wang, Z., Mariet, Z., Hu, H., et~al.
\newblock Plex: Towards reliability using pretrained large model extensions.
\newblock \emph{arXiv preprint arXiv:2207.07411}, 2022.

\bibitem[Valdenegro-Toro \& Mori(2022)Valdenegro-Toro and Mori]{valdenegro2022deeper}
Valdenegro-Toro, M. and Mori, D.~S.
\newblock A deeper look into aleatoric and epistemic uncertainty disentanglement.
\newblock In \emph{Computer Vision and Pattern Recognition Workshops (CVPRW)}, 2022.

\bibitem[van~der Maaten \& Hinton(2008)van~der Maaten and Hinton]{JMLR:v9:vandermaaten08a}
van~der Maaten, L. and Hinton, G.
\newblock Visualizing data using {t-SNE}.
\newblock \emph{Journal of Machine Learning Research (JMLR)}, 9\penalty0 (86):\penalty0 2579--2605, 2008.

\bibitem[Wah et~al.(2011)Wah, Branson, Welinder, Perona, and Belongie]{cub200}
Wah, C., Branson, S., Welinder, P., Perona, P., and Belongie, S.
\newblock The {Caltech-UCSD} birds-200-2011 dataset.
\newblock Technical Report CNS-TR-2011-001, California Institute of Technology, 2011.

\bibitem[Wightman(2019)]{rw2019timm}
Wightman, R.
\newblock {PyTorch} image models, 2019.

\bibitem[Wimmer et~al.(2023)Wimmer, Sale, Hofman, Bischl, and H{\"u}llermeier]{wimmer2023quantifying}
Wimmer, L., Sale, Y., Hofman, P., Bischl, B., and H{\"u}llermeier, E.
\newblock Quantifying aleatoric and epistemic uncertainty in machine learning: Are conditional entropy and mutual information appropriate measures?
\newblock In \emph{Uncertainty in Artificial Intelligence (UAI)}, 2023.

\bibitem[{Xiao} et~al.(2010){Xiao}, {Hays}, {Ehinger}, {Oliva}, and {Torralba}]{Xiao2010}
{Xiao}, J., {Hays}, J., {Ehinger}, K.~A., {Oliva}, A., and {Torralba}, A.
\newblock Sun database: Large-scale scene recognition from abbey to zoo.
\newblock In \emph{Conference on Computer Vision and Pattern Recognition (CVPR)}, 2010.

\bibitem[Yoo \& Kweon(2019)Yoo and Kweon]{Yoo_2019_CVPR}
Yoo, D. and Kweon, I.~S.
\newblock Learning loss for active learning.
\newblock In \emph{Conference on Computer Vision and Pattern Recognition (CVPR)}, 2019.

\bibitem[Yun et~al.(2019)Yun, Han, Oh, Chun, Choe, and Yoo]{yun2019cutmix}
Yun, S., Han, D., Oh, S.~J., Chun, S., Choe, J., and Yoo, Y.
\newblock Cutmix: {R}egularization strategy to train strong classifiers with localizable features.
\newblock In \emph{International Conference on Computer Vision (ICCV)}, 2019.

\bibitem[Zhai et~al.(2020)Zhai, Puigcerver, Kolesnikov, Ruyssen, Riquelme, Lucic, Djolonga, Pinto, Neumann, Dosovitskiy, Beyer, Bachem, Tschannen, Michalski, Bousquet, Gelly, and Houlsby]{zhai2020largescale}
Zhai, X., Puigcerver, J., Kolesnikov, A., Ruyssen, P., Riquelme, C., Lucic, M., Djolonga, J., Pinto, A.~S., Neumann, M., Dosovitskiy, A., Beyer, L., Bachem, O., Tschannen, M., Michalski, M., Bousquet, O., Gelly, S., and Houlsby, N.
\newblock A large-scale study of representation learning with the visual task adaptation benchmark, 2020.

\bibitem[Zhang et~al.(2018)Zhang, Cisse, Dauphin, and Lopez-Paz]{zhang2018mixup}
Zhang, H., Cisse, M., Dauphin, Y.~N., and Lopez-Paz, D.
\newblock Mixup: {B}eyond empirical risk minimization.
\newblock In \emph{International Conference on Learning Representations (ICLR)}, 2018.

\end{thebibliography}
\bibliographystyle{icml2024}

\newpage
\appendix
\onecolumn

\section{Training details} \label{app:details}

\textbf{Architecture.} With $d_e$ denoting the dimensionality of the (flattened) embeddings $e(x)$ of each ViT size, our pretrained uncertainty module has the following size across all ViT sizes: \texttt{Linear($d_e$, 512), LeakyReLU (negative slope 0.01), Linear(512, 512), LeakyReLU (negative slope 0.01), Linear(512, 1), Softplus($\beta$=1, threshold=20)}. The softplus in the end is to ensure that all uncertainties are strictly positive. This could be dropped since our uncertainties are scale free, but we added it for convenience of interpretation.

\textbf{Optimizer.} We train on ImageNet-21k-W for 460 episodes of 200,000 images, corresponding to roughly seven full epochs. We use a cosine learning rate scheduler that warms up the learning rate from 0.0001 to 0.0028 for 25 episodes and then decays it down to 1e-8 for the remaining episodes. We use an AdamW \cite{loshchilov2017decoupled} optimizer with $\beta_1=0.8$ and $\beta_2=0.95$. We apply weight decay of strength $0.0001$. These settings are constant for all experiments, without any hyperparameter tuning. 

\textbf{Augmentations.} We use the torchvision \cite{torchvision2016} augmentations that \texttt{timm} applies by default. Inter alia, all images are cropped to 224x224 pixels. We first apply a \texttt{RandomResizedCropAndInterpolation(size=(224, 224), scale=(0.08, 1.0), ratio=(0.75, 1.3333), interpolation=bilinear bicubic)}, and then randomly add \texttt{RandomHorizontalFlip($p$=0.5)} and with $p=0.4$ a \texttt{ColorJitter(brightness=[0.6, 1.4], contrast=[0.6, 1.4], saturation=[0.6, 1.4], hue=None)}.

\section{Transfer analysis} \label{app:generalize}
In this section, we analyze which datasets our pretrained uncertainties transfer to. We hypothesize that they behave similarly to the classifier head for ImageNet-21k. To test this, we use the entropy of the 21k class predictions of the classifier head as uncertainties and test its R-AUROC. We find that our pretrained uncertainties achieve a similar performance on all datasets. This indicates that pretrained uncertainties works on datasets similar enough to ImageNet-21k that its classifier is also informative. Note that this classifier method is not applicable to provide pretrained uncertainties in practice since it requires maintaining a heavy classifier head (17M parameters for ViT-Base), violating principle (iv), is not scalable to datasets with more classes, violating principle (v), and is not available outside ImageNet-21k classification, violating principle (iii). 

\begin{table}[h]
\centering
\footnotesize
\begin{tabular}{lcc}
\toprule
     Dataset &  Pretrained Uncertainties & 21k Classifier Entropy \\
     \cmidrule(lr){1-3}
     ImageNet-21k & \res{0.791}{0.001} & 0.798 \\
     Caltech 101 & \res{0.758}{0.006} & 0.808 \\ 
     Oxford Pets & \res{0.740}{0.008} & 0.724 \\ 
     CIFAR 10 & \res{0.739}{0.002} & 0.716 \\ 
     CIFAR 100 & \res{0.706}{0.002} & 0.696 \\ 
     SUN & \res{0.691}{0.002} & 0.697 \\ 
     Oxford Flowers & \res{0.659}{0.009} & 0.679 \\ 
     Describable Textures & \res{0.649}{0.006} & 0.610 \\ 
     CUB 200 & \res{0.626}{0.008} & 0.608 \\ 
     Stanford Online Products & \res{0.607}{0.001} & 0.591 \\ 
     CARS 196 & \res{0.589}{0.003} & 0.554 \\ 
     Treeversity & \res{0.560}{0.003} & 0.565 \\ 
     SVHN & \res{0.495}{0.005} & 0.524 \\ 
     \bottomrule
\end{tabular}
\caption{Pretrained uncertainties perform similarly to using the entropy of the ImageNet-21k classifier head as uncertainty estimate (note that this is impractical due to its size and violating the first principles in \cref{sec:principles}). This implies that pretrained uncertainties cover roughly the classes that ImageNet-21k also covers.}
\end{table}

\section{Simple beats Complex: Negative Results} \label{app:negative}

We test multiple adjustments to our loss, architecture and optimizer in \cref{tab:negative}. We report the average R-AUROC on the unseen datasets CUB, CARS, and SOP as in the URL protocol, evaluated for five seeds on a ViT-Base pretrained on ImageNet-21k-W. 

The first change regards the loss function. Currently, when comparing two images, it always requires one image to have a pretrained uncertainty of at least $0.1$ larger than the other one. In the formula below, we add an 'approximately equal' category, such that images whose ground-truth loss is within a leeway $l$ are not required to have different loss values:
\begin{align}
    \mathbbm{1}_\mathcal{L} := \begin{cases}
        +1 & \text{, if } \mathcal{L}_\text{task}^\text{det.}(y_1, f(x_1)) > l + \mathcal{L}_\text{task}^\text{det.}(y_2, f(x_2)) \\
        -1 & \text{, if } \mathcal{L}_\text{task}^\text{det.}(y_1, f(x_1)) + l < \mathcal{L}_\text{task}^\text{det.}(y_2, f(x_2)) \\
        0 & \text{, else} 
    \end{cases} \,.
\end{align}
However, at several values of the allowed leeway $l$, this does not change the performance outside the margin of error of the baseline method (\res{0.608}{0.004}). 

Second, we change the size of the uncertainty head MLP. By default it has 2 hidden layers of width 512. We either shrink it to 1 hidden layer with width 256 or enlargen it to 3 hidden layers of width 1024. While there is a slight trend favoring smaller heads, it does not exceed the margin of chance.

Third, we briefly experimented with initializing the uncertainty module with zero-weights. However, this failed to train at all, which is theoretically expected.

Fourth, we add strong augmentations that add different types of aleatoric uncertainty to half of the train dataset. None of these increase the performance, with some even deteriorating it. While this might seem counterintuitive, we presume such artificial sources of uncertainty do not reflect the uncertainties occuring on real images. 

Last, we experiment with optimizers other than our default AdamW with cosine learning rate scheduler. While Lion collapses after less than one epoch, SGD performs slightly better than the baseline. However, it might be a false positive, especially taking multiple testing into account. Indeed, the reason we did not select SGD for the main paper is that it did not systematically outperform AdamW during our preliminary experiments on the validation splits with less seeds. The test splits were held strictly secret until the writing of the paper. We suggest future researchers to experiment with replacing their advanced optimizers by SGD.

\begin{table}[h]
\centering
\footnotesize
    \begin{tabular}{lr}
    \toprule
         Method & R-AUROC \\
    \cmidrule(lr){1-2}
         Default & \res{0.608}{0.004} \\ 
    \cmidrule(lr){1-2}
         Softened loss ($l=0.001$) & \res{0.607}{0.006} \\
         Softened loss ($l=0.01$) & \res{0.607}{0.005}\\
         Softened loss ($l=0.1$) & \res{0.609}{0.005} \\
    \cmidrule(lr){1-2}
        Smaller uncertainty module & \res{0.611}{0.002}\\
        Larger uncertainty module & \res{0.606}{0.004}\\
    \cmidrule(lr){1-2}
        Initialize uncertainty module with zero & \res{0.500}{0.000}\\
    \cmidrule(lr){1-2}
        AugMix \cite{hendrycks2020augmix} for 50\% of train data & \res{0.606}{0.005}\\
        CutMix \cite{yun2019cutmix} for 50\% of train data & \res{0.575}{0.002}\\
        MixUp \cite{zhang2018mixup} for 50\% of train data & \res{0.552}{0.009} \\
        Blurred images for 50\% of train data & \res{0.609}{0.003} \\
        Small crops for 50\% of train data & \res{0.610}{0.004}\\
        Box overlay for 50\% of train data & \res{0.606}{0.001}\\
    \cmidrule(lr){1-2}
        Adam optimizer \cite{kingma2014adam} & \res{0.609}{0.002}\\
        Lion optimizer \cite{chen2023symbolic} & \res{0.500}{0.000}\\
        SGD optimizer & \res{0.614}{0.005} \\ 
        Step learning rate scheduler & \res{0.607}{0.003}\\
    \bottomrule
    \end{tabular}
    \caption{No change to loss, architecture, optimizer, or data augmentation improves the performance. R-AUROC averaged across CUB, CARS, and SOP as in the URL protocol, for five seeds on a ViT-Base.}
    \label{tab:negative}
\end{table}

\end{document}